\documentclass{article}

\usepackage{arxiv}

\usepackage[utf8]{inputenc} 
\usepackage[T1]{fontenc}    
\usepackage{hyperref}       
\usepackage{url}            
\usepackage{booktabs}       
\usepackage{amsfonts}       
\usepackage{nicefrac}       
\usepackage{microtype}      
\usepackage{graphicx}
\usepackage{natbib}
\usepackage{doi}

\title{Just Another Method to Compute MTTF from Continuous Time Markov Chain Models}

\author{ \href{https://orcid.org/
0000-0003-0952-0689}{\includegraphics[scale=0.06]{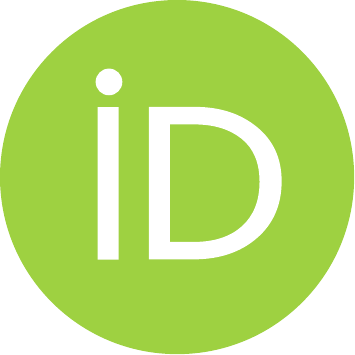}\hspace{1mm}Eduardo M. Vasconcelos} \\
	Academic Department of Superior Courses\\
	Federal Institute of Pernambuco\\
	Recife, PE 50740-540 \\
	\texttt{eduardo.vasconcelos@recife.ifpe.edu.br} \\
	\\
}

\hypersetup{
pdftitle={Just Another Method to Compute MTTF from Continuous Time Markov Chains},
pdfsubject={q-bio.NC, q-bio.QM},
pdfauthor={Eduardo M. Vasconcelos},
pdfkeywords={Meantime to Failure, Continuous Time Markov Chain, 
Stochastic Process, Business Processes},
}

\begin{document}
\maketitle

\begin{abstract}
	The Meantime to Failure is a statistic used to determine how much time a system spends to enter one of its absorption states. This statistic can be used in most areas of knowledge. In engineering, for example, can be used as a measure of equipment reliability, and in business, as a measure of processes performance. This work presents a method to obtain the Meantime to Failure from a Continuous Time Markov Chain models. The method is intuitive and is simpler to be implemented, since, it consists of solving a system of linear equations.
\end{abstract}


\section{Introduction}
\label{sec:introduction}

The Meantime To Failure (MTTF) is a statistic used for system analysis in several knowledge areas. This value represents the average time to the system enters into one of the possible states of fault, without considering system repairs. Although MTTF be considered to analyze systems with fault states, it also can be used to perform analysis on processes, since it can be used to represent the meantime to one process finishes, given that, processes can be represented by state machine models.

This work presents a method to compute MTTF from Continuous Time Markov Chain (CTMC) models. There are no arguments that demonstrate that this method performs better than other methods, but this method has a simpler implementation and is intuitive. This method also allows computing the absorption probabilities and the average holding time of each state without additional steps.

\section{Calculating the Meantime to Failure}
\label{sec:headings}

This section presents a method to compute the MTTF from a CTMC model with absorption states. This method consists in generating a secondary model from the original, adding on it unitary repairs.

To extract the MTTF from a CTMC model, it´s necessary that this model accomplishes two conditions: (i) the model need to have a initial state $s_\iota$; (ii) the model need to have at last one absorption state. Let´s consider a CTMC $Z=\{S,T\}$, with $Z$ containing a set of states $S = \{s_1, s_2,...,s_n\}$ and a set of transitions $T=\{t_{i,j}\}$ with $t_{i,j} = 1$ if there is a transition between states $s_i$ and $s_j$, and $t_{i,j}=0$, otherwise. A state is considered an absorption state $s_f$ if $t_{f,k}=0, \forall k$. The occurrence rate of a transition is $q_{i,j}>0$ if $t_{i,j}=1$ and $q_{i,j}=0$ if $t_{i,j}=0$, with $q_{i,i}= -\sum{q_{i,\beta}}$ and $\beta \neq i$.

\begin{figure}
	\centering
	\includegraphics[width=10cm]{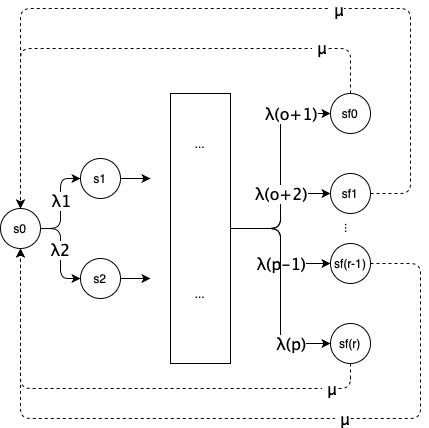}
	\caption{A General View of $Z'$}
	\label{fig:fig1}
\end{figure}

To compute the MTTF we need to create a secondary model $Z' = \{[S_A,S_F],[T, T']\}$, with $S_A$ the set of reachable states and the initial state and $S_F$ the set of fault states. After creating $Z'$ we have to add repair transitions $T'$ between states $S_F$ and $s_\iota$ with $q_{f,\iota}=\mu$ as showed on Figure \ref{fig:fig1}. The figure presents $Z$ with the addiction of repair transitions $T'$ with rate $\mu$, which are represented by dashed arrows.

From $Z'$, which corresponds to $Z$ with addiction of $T'$, being $Z'$ a closed chain, it´s possible to compute the steady-state availability $A'$. The process of computing the availability can be seen in \citep{10.5555/289350}. To compute the MTTF from $Z'$ and $A'$, we can use the expression \ref{eq:eq1}.

\begin{equation} \label{eq:eq1}
	A'=\frac{MTTF}{MTTF + MTTR}
\end{equation}

Where MTTR corresponds to the Meantime to repair the system. As transitions $T'$ were artificially added on $Z'$, the MTTR can be computed according to equation \ref{eq:eq2}.

\begin{equation} \label{eq:eq2}
	MTTR = \sum{P[sf_{u}]  \mu^{-1} = \mu^{-1}} \sum{P[sf_{u}]}
\end{equation}

Where $P[.]$ is the probability of a fault state be reached at the end of the process and $sf_u \in S'_F$. As $\sum{P[sf_{u}]} = 1$, given that every fault states can be reached, $MTTR = \mu^{-1}$. So, we have:

\begin{equation} \label{eq:eq3}
	MTTF = -\frac{A'  \mu^{-1}}{A' - 1} = \frac{A'  \mu^{-1}}{U'}
\end{equation}

Where $U'$ is the unavailability of the system.

Although the MTTF of equation \ref{eq:eq3} be computed from the availability of $Z'$, this value corresponds to the MTTF of $Z$, since the set of states and transitions of $Z'$ are the same as the original. The method to calculate the MTTF is easier to be implemented than the methods presented in \citep{trivedi_bobbio_2017} since it´s just necessary to solve a system of linear equations presented on equation \ref{eq:eq4} \citep{DBLP:journals/corr/Vasconcelos17}.

\begin{equation} \label{eq:eq4}
	\pi'_{i} = \frac{\sum{\pi'_j  q_{j,i}  t_{j,i}}}{\sum{q_{i,k}  t_{i,k}}}
\end{equation}

From the MTTF and $A'$ it is possible to compute other values of interest to the analysis of $Z$. As $A' = \sum{\pi'_a}$ with $\pi'_a$ representing the steady probability of state $s_a \in S_A$, and $U'=1-A'=\sum{\pi'_f}$ with $\pi'_f$ as the steady probability of $s_f \in S_F$, we have the following values: (i) $\tau_a = MTTF. (\pi'_a A'^{-1})$, (ii) $\rho_f = \pi'_f  U'^{-1}$. Where $\tau_a$ corresponding to the average holding time of state $s_a$, $\rho_f$ corresponding to the probability of $s_f$ be reached.

The value $\tau_a$ is obtained from a proportion of MTTF since the MTTF is the sum of the average holding time of states $S_A$ as showed in \citep{trivedi_bobbio_2017}.

\section{Conclusions}

This work presented a method to compute the Meantime to Failure (MTTF) from Continuous Time Markov Chain Models with absorption states. The method consists of creating a second model and adding repair transitions to it. From the secondary model, it is possible to calculate the steady-state availability and, from this value, calculate the MTTF. Beyond the MTTF, it´s possible to compute probabilities of the process reaching the fault states, as well as the states holding time.

\bibliographystyle{unsrtnat}
\bibliography{references}

\end{document}